\documentclass{article}
\usepackage{ijcai26}

\usepackage{times}
\usepackage{soul}
\usepackage{url}
\usepackage[hidelinks]{hyperref}
\usepackage[utf8]{inputenc}
\usepackage[small]{caption}
\usepackage{graphicx}
\usepackage{amsmath}
\usepackage{amsthm}
\usepackage{booktabs}
\usepackage{algorithm}
\usepackage{algorithmic}
\usepackage[switch]{lineno}
\usepackage{subdepth}

\usepackage{tikz}

\title{PEACE 2.0: \\ Grounded Explanations and Counter-Speech for Combating Hate Expressions}

\author{
Greta Damo$^{1, a}$
\and
Stéphane Petiot$^{2, b}$\and
Elena Cabrio$^{1, a}$\And
Serena Villata$^{1, a}$\\
\affiliations
$^1$Université Côte d'Azur, CNRS, Inria, I3S, France\\
$^2$Université Côte d’Azur, Institut 3IA Côte d’Azur, Techpool, France\\
\emails
$^a$\{firstname.surname\}@univ-cotedazur.fr,
$^b$stephane.petiot@inria.fr
}

\begin{document}

\maketitle

\begin{abstract}

The increasing volume of hate speech on online platforms poses significant societal challenges. While the Natural Language Processing community has developed effective methods to automatically detect the presence of hate speech, responses to it, called counter-speech, are still an open challenge. We present PEACE 2.0, a novel tool that, besides analysing and explaining why a message is considered hateful or not, also generates a response to it. More specifically, PEACE 2.0 has three main new functionalities: leveraging a Retrieval-Augmented Generation (RAG) pipeline \textit{i)} to ground HS explanations into evidence and facts, \textit{ii)} to automatically generate evidence-grounded counter-speech, and \textit{iii)} exploring the characteristics of counter-speech replies. 
By integrating these capabilities, PEACE 2.0 enables in-depth analysis and response generation for both explicit and implicit hateful messages. 





\end{abstract}

\section{Introduction}

The growing volume of hate speech (HS) on social media represents a major societal challenge, requiring automated tools that go beyond surface-level detection. Hate speech can appear in explicit, implicit, or subtle forms, with the latter relying on coded, indirect, or context-dependent language that is particularly difficult to identify and interpret \cite{ocampo2023depth}. While advances in Natural Language Processing (NLP) and Generation (NLG) have improved detection of explicit content, understanding and addressing nuanced forms of hate remains an open problem.
To address this challenge, 
the tool PEACE was introduced (Providing Explanations and Analysis for Combating Hate Expressions) \cite{damo2024peace}, a web-based system designed to support exploration, detection, and explanation of explicit, and implicit HS. PEACE enables content moderators and researchers to analyze hateful messages, inspect model predictions, and obtain natural language explanations clarifying why a message is considered hateful.
In this paper, we present PEACE 2.0, an extended version that moves beyond analysis and explanation to support actionable responses through automatic counter-speech (CS) generation. Counter-speech challenges hateful messages with factual information or alternative perspectives, offering a constructive alternative to content removal or censorship \cite{benesch2014dangerous}. However, generating effective CS is particularly challenging for implicit hate, where responses must be both informative and carefully framed.
PEACE 2.0 introduces three key new functionalities to address these challenges:
\textbf{1. Knowledge-grounded counter-speech generation.} A Retrieval-Augmented Generation (RAG) pipeline, as in 
\cite{damo2025beating}, retrieves evidence from authoritative human rights sources and conditions counter-speech on this information. Users can explicitly compare responses generated with and without retrieval, highlighting the impact of knowledge grounding on factuality, relevance, and effectiveness.
\textbf{2. Evidence-grounded explanations for hate speech classification.} Using the same RAG mechanism, PEACE 2.0 generates explanations that justify the predictions of a fine-tuned BERT classifier, grounding label decisions in retrieved evidence to improve transparency and interpretability.
\textbf{3. Visual analytics for counter-speech exploration.} Interactive tools allow analysis of numerous counter-speech datasets supporting experimental evaluation of counter-speech performance on explicit versus implicit hate speech and offering insights into the role of retrieval-based grounding.

\noindent \textbf{Related Work.} Existing systems address HS detection, visualization, monitoring, or CS generation separately. Tools such as RECAST, MUDES, MUTED, IFAN, CRYPTEXT, TweetNLP, and McMillan-Major et al.’s framework provide detection and analysis interfaces \cite{recast,mudes,muted,ifan,cryptext,tweetnlp,mcmillan-major}, while dashboards like the Indonesian HS monitoring system \cite{nlpindonesia} support large-scale tracking. In the CS domain, work is limited; CounterHelp \cite{counterhelp} leverages large language models for context-sensitive responses. 
However, prior systems treat these components in isolation, focus mainly on explicit hate, and do not systematically compare knowledge-grounded and non-grounded generation.

Overall, PEACE 2.0 offers a unified, interactive platform for hate speech exploration, detection, explanation, and response generation, bridging analytical insights and practical mitigation. To our knowledge, it is the only online tool providing in-depth analysis of both explicit and implicit hate speech together with knowledge-grounded CS generation.


\section{PEACE 2.0 Main Functionalities}

In this section, we report the main features of the original demo, and the new functionalities of PEACE 2.0\footnote{The demo video is available \href{https://youtu.be/5nZOP5HfLc8}{at this link}. We also provide a public API built on Python, Flask, and JavaScript \href{https://gitlab.inria.fr/nocampo/peace}{here}.}.

\subsection{Data Exploration \& Visualization}
This module offers interactive visualizations for exploring HS and CS datasets. Users can switch between the two views.

\noindent \textbf{Hate Speech.} The hate speech view covers implicit hate datasets used in PEACE: the Implicit Hate Corpus (IHC) \cite{ihc}, Implicit and Subtle Hate (ISHate) \cite{ocampo2023depth}, TOXIGEN \cite{toxigen}, DynaHate (DYNA) \cite{dyna}, and Social Bias Inference Corpus (SBIC) \cite{sbic}. Messages are organized by hatefulness, implicitness, and target group, with sanitized labels for consistency, that users can filter. 
Visualizations include: \textit{Sankey diagrams} linking target groups and HS categories (explicit vs. implicit) with topic distributions derived via Latent Dirichlet Allocation (LDA); \textit{Word Clouds} displaying the most frequent lexical items within selected attributes; and \textit{Target Frequency} charts displaying attacked groups across datasets and implicitness levels.

\noindent\textbf{Counter-speech.} In PEACE 2.0, the same visual framework is extended to CS datasets, including CONAN \cite{conan}, Multitarget-CONAN \cite{mtconan}, Twitter and YouTube datasets \cite{mathew2020twitter,albanyan-twitter,mathew2019thou}, knowledge-grounded human expert datasets \cite{bonaldi2025first,towards-kg}, and RAG-generated responses from multiple LLMs and retrieval strategies \cite{damo2025beating}.
Labels follow the same sanitization scheme as PEACE. 
\textit{Sankey diagrams} connect targets, counter-speech sources (expert, user, RAG, No-RAG), and LDA topics; \textit{Word Clouds} highlight frequent terms within filtered messages, and \textit{Frequency} charts display CS volume per target and source, and the distribution of different CS strategies across targets. 

\subsection{Data Augmentation}
This module generates adversarial examples by modifying messages while preserving their implicit hateful meaning. The objective is to augment data for implicit HS by altering surface-level elements without changing the underlying stance. Following \cite{ocampo2023depth}, 
implemented strategies include: named entity replacement; adjustment of scalar adverbs; addition of adverbial modifiers; adjective synonym substitution; replacement of domain-specific expressions with semantically similar variants; Easy Data Augmentation (random replacement, insertion, swap, deletion); and back-translation. Users can apply these methods to custom inputs.

\subsection{Hate Speech Detection and Explanation}

This module allows users to input custom messages for binary hate speech classification. The system outputs the predicted label with its confidence score. In addition, PEACE 2.0 generates concise, human-readable explanations conditioned on the message, predicted label, and probability, helping users understand the rationale behind the decision.
PEACE 2.0 further introduces an optional RAG pipeline from \cite{damo2025beating}, to produce evidence-grounded explanations. When enabled, the system retrieves relevant passages from an authoritative knowledge base, summarizes them into factual context, and integrates this information into the LLM prompt. Users can choose whether to use RAG and select the underlying LLM. For transparency, the system also shows the retrieved paragraphs together with their similarity score. 


\subsection{Knowledge-Grounded CS Generation}
Building on the RAG pipeline, a central innovation in PEACE 2.0 is evidence-grounded CS generation based on a curated human rights knowledge base. 
Users can input a hateful message and generate counter-speech responses with or without knowledge grounding, enabling direct comparison of evidence on outputs in terms of relevance, factuality, and persuasiveness. Multiple LLMs are supported, offering flexibility in style and quality.
The CS generation pipeline consists of three steps. First, the input message is encoded using the BGE-M3 sentence transformer, and FAISS performs inner-product similarity search over precomputed paragraph embeddings to retrieve the top-$3$ most relevant evidence passages, with deduplication applied. 
Second, the retrieved passages are concatenated and summarized into a concise summary using the selected LLM. 
Finally, the original message and the evidence summary are provided to the same LLM to generate a respectful and persuasive CS response suitable for social media. 
Users may also disable retrieval to compare RAG and non-RAG outputs. For transparency, the system also shows the retrieved paragraphs together with their similarity score.

\noindent \textbf{Available models.} For detection, the demo uses a BERT classifier fine-tuned on the ISHate training set. 
For explanation and CS generation, PEACE 2.0 supports open-source LLMs: Mistral (\texttt{Mistral-7B-Instruct-v0.3}), LLaMa (\texttt{Llama-3.1-8B-Instruct}), and CommandR (\texttt{c4ai-command-r7b-12-2024}). The knowledge base comprises 32,792 documents from the United Nations Digital Library, Eur-Lex, and the European Agency for Fundamental Rights (from 2000 to 2025), totaling 3,173,630 tokenized paragraphs. 


\section{Experiments and Results}

\subsection{Experimental Setting}

We hypothesize that RAG-grounded explanations and CS will outperform non-RAG generations for both explicit and implicit HS, demonstrating the value of evidence-grounded generation. Specifically, we test: 
\textbf{H1}: RAG outputs are better overall and more informative for both explanations and CS.
\textbf{H2}: RAG improves persuasiveness;
\textbf{H3}: RAG improves outputs for both implicit and explicit cases.
To evaluate this, 
from the implicit HS datasets (IHC, ISHate, TOXIGEN, DYNA, SBIC), we randomly sample 20 messages per dataset, evenly split between explicit and implicit, resulting in 100 HS examples. For each message, we generate explanations with and without RAG (100 RAG, 100 non-RAG). The same procedure is applied to CS, producing 200 responses. We assess all generations through both automatic and human evaluation.

\noindent \textbf{Human Evaluation.} We evaluate a subset of 100 explanations and 100 CS responses. For each task, 50 explicit and 50 implicit (25 RAG, 25 non-RAG) cases are sampled, resulting in 200 human-evaluated instances overall. Each example is assessed by three trained annotators. 
Evaluation criteria for explanations follow PEACE, 
while CS metrics are adapted from \cite{zheng2023,bonaldi2024safer,damo2025effectiveness}. Both explanations and CS are rated on: \textbf{Fluency (F)} (grammatical correctness), \textbf{Informativeness (I)} (relevant contextual or factual content), \textbf{Persuasiveness (P)} (ability to convince the hater and foster empathy in bystanders), \textbf{Soundness (SO)} (logical coherence), and \textbf{Specificity (SP)} (direct engagement with the HS and target). All dimensions are rated on a 1-5 Likert scale (5 = highest).

\noindent \textbf{Automatic metrics.} We compute automatic measures of linguistic quality, diversity, and faithfulness. These include \textit{Distinct-3} for lexical diversity; \textit{Semantic Similarity} (Sentence-BERT) between HS and generated explanations or counter-speech; and \textit{Perplexity} as a proxy for fluency. For RAG outputs, we also measure \textit{faithfulness} via similarity between generations and retrieved evidence. Finally, \textit{NLI-based metrics} (entailment and contradiction with \texttt{roberta-large-nli}) assess whether outputs appropriately address the original HS and remain consistent with the supporting evidence.


\begin{table}[!ht]
\centering
\small
\begin{tabular}{lcccc}
\toprule
\textbf{Metric} & \textbf{$Exp_{\text{RAG}}$} & \textbf{$Exp_{\text{NoRAG}}$} & \textbf{$Imp_{\text{RAG}}$} & \textbf{$Imp_{\text{NoRAG}}$} \\
\midrule
\multicolumn{5}{c}{\textbf{Explanations}} \\
\midrule
F & 5.00 & 5.00 & 5.00 & 5.00 \\
SO & 4.88 & 4.56 & 4.80 & 4.58 \\
I & 4.38 & 2.84 & 4.64 & 2.72 \\
SP & 4.86 & 3.78 & 4.88 & 4.40 \\
P & 4.68 & 3.52 & 4.72 & 3.86 \\
\textbf{Overall} & \hl{\textbf{4.76}} & \textbf{3.94} & \hl{\textbf{4.81}} & \textbf{4.11} \\
\midrule
\multicolumn{5}{c}{\textbf{Counter-speech}} \\
\midrule
F & 5.00 & 5.00 & 5.00 & 5.00 \\
SO & 4.82 & 3.92 & 4.88 & 4.52 \\
I & 4.66 & 2.52 & 4.80 & 2.86 \\
SP & 4.90 & 2.98 & 4.90 & 3.32 \\
P & 4.68 & 2.64 & 4.94 & 3.22 \\
\textbf{Overall} & \hl{\textbf{4.81}} & \textbf{3.41} & \hl{\textbf{4.90}} & \textbf{3.78} \\
\bottomrule
\end{tabular}
\caption{Mean human ratings for RAG vs. No-RAG outputs.}
\label{tab:human_eval}
\end{table}

\begin{table}[!ht]
\centering
\footnotesize
\begin{tabular}{lcccc}
\toprule
\textbf{Metric} & 
\textbf{$Exp_{\scriptstyle \text{RAG}}$} & 
\textbf{$Exp_{\scriptstyle \text{NoRAG}}$} & 
\textbf{$Imp_{\scriptstyle \text{RAG}}$} & 
\textbf{$Imp_{\scriptstyle \text{NoRAG}}$} \\
\midrule
\multicolumn{5}{c}{\textbf{Explanations}} \\
\midrule
Sem. Sim. & \textbf{0.57} & 0.49 & \textbf{0.56} & 0.47 \\
Faithfulness & 0.57 & - & 0.56 & - \\
Perplexity & \textbf{25.66} & 37.38 & \textbf{25.43} & 37.21 \\
Distinct-3 & 0.98 & \textbf{0.99} & 0.97 & \textbf{0.99} \\
Hate-Ent. & \textbf{0.18} & 0.08 & \textbf{0.12} & 0.08 \\
Ev.-Contr. & 0.05 & - & 0.05 & - \\
Ev.-Ent. & 0.26 & - & 0.21 & - \\
\midrule
\multicolumn{5}{c}{\textbf{Counter-speech}} \\
\midrule
Sem. Sim. & \textbf{0.51} & 0.45 & \textbf{0.50} & 0.41 \\
Faithfulness & 0.65 & - & 0.61 & - \\
Perplexity & \textbf{14.47} & 21.74 & \textbf{14.36} & 22.91 \\
Distinct-3 & 0.99 & \textbf{1.00} & 0.99 & \textbf{1.00} \\
Hate Ent. & \textbf{0.04} & 0.03 & \textbf{0.09} & 0.05 \\
Ev.Contr. & 0.05 & - & 0.03 & - \\
Ev. Ent. & 0.18 & - & 0.15 & - \\
\bottomrule
\end{tabular}
\caption{Automatic metrics results. Abbreviations are for: Semantic Similarity (Sem. Sim.), Entailment (Ent.), Contradiction (Contr.), Evidence (Ev.). Best results are in bold.} 
\label{tab:auto_metrics}
\end{table}

\subsection{Results}

\noindent \textbf{Human evaluation.} From Table \ref{tab:human_eval}, RAG-generated outputs consistently outperform non-RAG outputs across all metrics. Supporting H1, \textit{RAG outputs are significantly more informative}, particularly for implicit content (Explanation-Imp.: 4.64 vs. 2.72; Counter-speech-Imp.: 4.80 vs. 2.86), \textit{and of better quality} (see highlighted results). In line with H2, \textit{RAG also improves persuasiveness} for both explicit and implicit hate (e.g., Explanation-Exp.: 4.68 vs. 3.52; Counter-speech-Exp.: 4.68 vs. 2.64). Consistent with H3, \textit{gains are larger for both implicit and explicit cases}, highlighting the benefit of retrieval in subtle contexts. Fluency and Soundness remain comparable across RAG and non-RAG outputs, indicating that these improvements do not compromise readability or coherence. Wilcoxon signed-rank tests confirm these differences are statistically significant (p $<$ 0.05). 
Inter-annotator agreement, measured using Krippendorff's $\alpha$, is substantial to perfect across dimensions ($k$ = 0.57-1).

\noindent \textbf{Automatic metrics.} From Table \ref{tab:auto_metrics}, we see that RAG-generated outputs consistently outperform No-RAG in both Explanation and CS tasks. They achieve higher semantic similarity, are faithful to retrieved evidence, have lower perplexity, indicating more informative and fluent outputs, while maintaining diversity (Distinct-3). NLI metrics further show that RAG outputs have higher Hate-Entailment and low Evidence-Contradiction, indicating that they appropriately address the content of the original HS without supporting it, and are aligned with retrieved content. Gains are for both implicit and explicit content, confirming that RAG improves informativeness, persuasiveness, and faithfulness across both tasks. Differences are statistically significant with Wilcoxon signed-rank tests.

\section{Conclusion}

By integrating retrieval, summarization, and generation within a single pipeline, the PEACE 2.0 tool  bridges analytical understanding and practical mitigation of both implicit and explicit hate speech messages, producing evidence-backed counter-speech that improves trustworthiness and effectiveness. The explanation and generation modules enhance the transparency of the abusive language classification and the counter-speech generation, making PEACE 2.0 a suitable tool for e-democracy applications with the aim to enhance inclusiveness and fairness. Future work includes the integration of adaptive retrieval strategies, dynamic knowledge base updates, and the inclusion of evaluation metrics to assess counter-speech quality. 




\bibliographystyle{named}
\bibliography{ijcai26}

\end{document}